\documentclass{ifacconf}

\usepackage{graphicx}      % include this line if your document contains figures
\usepackage{natbib}        % required for bibliography
\usepackage{amsmath}
\usepackage{amssymb}
\usepackage{graphicx}
\begin{document}
\begin{frontmatter}

\title{BOIS: Bayesian Optimization of Interconnected Systems} 

\thanks[footnoteinfo]{We acknowledge support from NSF-EFRI \#2132036, UW-Madison GERS program,  and the PPG Fellowship.}

\author{Leonardo D. Gonz\'alez and Victor M. Zavala}

\address{Department of Chemical and Biological Engineering, University of Wisconsin-Madison 
   Madison, WI 53706 USA}

\begin{abstract}                % Abstract of 50--100 words
Bayesian optimization (BO) has proven to be an effective paradigm for the global optimization of expensive-to-sample systems. One of the main advantages of BO is its use of Gaussian processes (GPs) to characterize model uncertainty which can be leveraged to guide the learning and search processes. However, BO typically treats systems as black-boxes and this limits the ability to exploit structural knowledge (e.g., physics and sparse interconnections). Composite functions of the form $f(x, y(x))$, wherein GP modeling is shifted from the performance function $f$ to an intermediate function $y$, offer an avenue for exploiting structural knowledge. However, the use of composite functions in a BO framework is complicated by the need to generate a probability density for $f$ from the Gaussian density of $y$ calculated by the GP (e.g., when $f$ is nonlinear it is not possible to obtain a closed-form expression). Previous work has handled this issue using sampling techniques; these are easy to implement and flexible but are computationally intensive. In this work, we introduce a new paradigm which allows for the efficient use of composite functions in BO; this uses adaptive linearizations of $f$ to obtain closed-form expressions for the statistical moments of the composite function. We show that this simple approach (which we call BOIS) enables the exploitation of structural knowledge, such as that arising in interconnected systems as well as systems that embed multiple GP models and combinations of physics and GP models. Using a chemical process optimization case study, we benchmark the effectiveness of BOIS against standard BO and sampling approaches. Our results indicate that BOIS achieves performance gains and accurately captures the statistics of composite functions.
\end{abstract}


\end{frontmatter}
%===============================================================================

\section{Introduction}

Optimization of natural and engineered systems (e.g., chemical processes, biological systems, materials) is often challenging due to incomplete physical knowledge or the high complexity of experiments and available physical models. As a result, it is necessary to device optimization procedures that effectively combine experimental and model data while mitigating complexity, see \cite{Conn:2009}. These procedures, often referred to as black-box optimization algorithms, treat the system as a black-box, $f(x)$, and use input/output observation data to direct their search for a solution. One of the most popular paradigms to have emerged in this setting is Bayesian optimization (BO), see \cite{Mockus:2012}. BO uses input/output data to generate a Gaussian process (GP) model that estimates system performance as well as model uncertainty. These estimates are used to construct an acquisition function (AF) that allows for the search to be tuned to emphasize sampling from either high performance (exploitation) or high uncertainty (exploration) regions; this is a key feature that makes BO an especially powerful global optimizer, see \cite{Shahriari:2016}.

While the black-box assumption substantially facilitates the implementation of BO (no prior knowledge about the system is needed), it ignores the fact that there is often some form of structural knowledge available (e.g., physics or sparse interconnectivity). For example, when dealing with physical systems, several components might be well-modeled and understood, while others might not. For those that are not, the fundamental principles governing their behavior (e.g., conservation laws, equilibrium, value constraints) are, at least qualitatively, understood. Additionally, sparse connectivity, which provides information on how different components affect each other, is also often known. As a result, the system of interest is usually not an unobservable black-box but rather a ``grey-box" that is partially observable with a known structure, see \cite{Astudillo:2021}. Several methods exist for performing grey-box BO; studies have shown that they are able to improve algorithm performance by effectively constraining the search space, resulting in lower sampling requirements and better solutions, see \cite{Lu:2021}; \cite{Raissi:2019}; \cite{Kandasamy:2017}. Of these approaches, the use of composite functions has proven to be one of the most widely used methods for incorporating preexisting knowledge into BO, see \cite{Astudillo:2019}.

A composite function expresses the system performance as $f(x, y(x))$ where $x$ are the system inputs, $f$ is a known scalar function, and $y$ is an unknown vector-valued function that describes the behavior of internal system components. This decomposition shifts the modeling task from estimating the performance function directly to estimating the values of $y$ which serve as inputs to $f(x, y(x))$. Additionally, because $f$ is now a known function, derivative information might be available to understand the effects of $x$ and $y$, see \cite{Urenholt:2019}. This approach also readily lends itself to the inclusion of constraints as these are often dependent on internal variables which can be captured by $y$ \cite{Paulson:2022}. Framing an optimization problem in this manner is therefore a fairly intuitive approach, especially in chemical engineering where the performance metric is usually an economic cost. For example, the cost equations for equipment, material streams, and utilities are often readily available and it is the parameters these equations rely on (compositions, flowrates, duties) that are unknown. Furthermore, traditional unit operations (heat exchangers, distillation columns, compressors) have significantly better mechanistic models available than those that tend to be more niche (bioreactors, non-equilibrium separators, solids-handling). It then makes sense to construct a composite function where the outer function, $f$, returns the price of the system based on the known cost equations while its inputs, $y$, are the mass and energy flows through the system and are estimated via either mechanistic or data-driven models. Constraints can then be incorporated using values estimated for $y$ to ensure that data-driven models obey fundamental physical laws as well as to enforce more traditional requirements such as product specifications, waste generation, utility consumption, and equipment sizing which are often important in process design. 

While setting up a composite function optimization problem seems fairly straightforward, implementing it in a BO setting is not a trivial task. As previously stated, one of the main advantages of BO is the inclusion of uncertainty estimates in the surrogate model, which allows for greater exploration of the design space when compared to a deterministic model, see \cite{boukouvala:2017}. However, when using a composite function, the GP models generated are of $y$ not $f$. Given that $f$ is the performance measure being optimized, it is necessary propagate the predicted uncertainty from $y(x)$ to $f(x, y(x))$ (i.e, the density of $f$ or desired summarizing statistics must be determined). A Gaussian density for $y(x)$ is directly obtained from the GP model; as a result, when $f$ is a linear model, we can make use of the closure of Gaussian random variables under linear operations to generate the density of $f(x, y(x))$ (which is also a Gaussian). When $f$ is nonlinear, however, a closed-form solution is not readily available, and this operation is usually carried out numerically using sampling methods like Monte Carlo (MC), see \cite{Balandat:2020}.

Given the increased functionality composite functions can provide to BO and the computational intensity of sampling methods, we propose the Bayesian Optimization of Interconnected Systems (BOIS) framework. BOIS provides a novel method for estimating the distribution of $f$ in a significantly more efficient manner than sampling by linearizing the performance function in the neighborhood of a $y(x)$ of interest. This allows us to construct a local Laplace approximation which we can use to generate closed-form expressions for the mean and uncertainty of $f(x, y(x))$. We test the performance of this approach by conducting extensive numerical experiments on a chemical process optimization case study and compare its performance to standard BO as well as MC-driven composite function BO. Our results illustrate that BOIS is able to outperform standard BO while also providing accurate estimates for the distribution of $f$ at a significantly lower computational cost than MC.

\footnotetext[1]{Corresponding author: Victor M. Zavala (e-mail: zavalatejeda@wisc.edu)}

\section{Background}

\subsection{Bayesian optimization}
\begin{figure}[tb]
	\centering
	\includegraphics[width=0.49\textwidth]{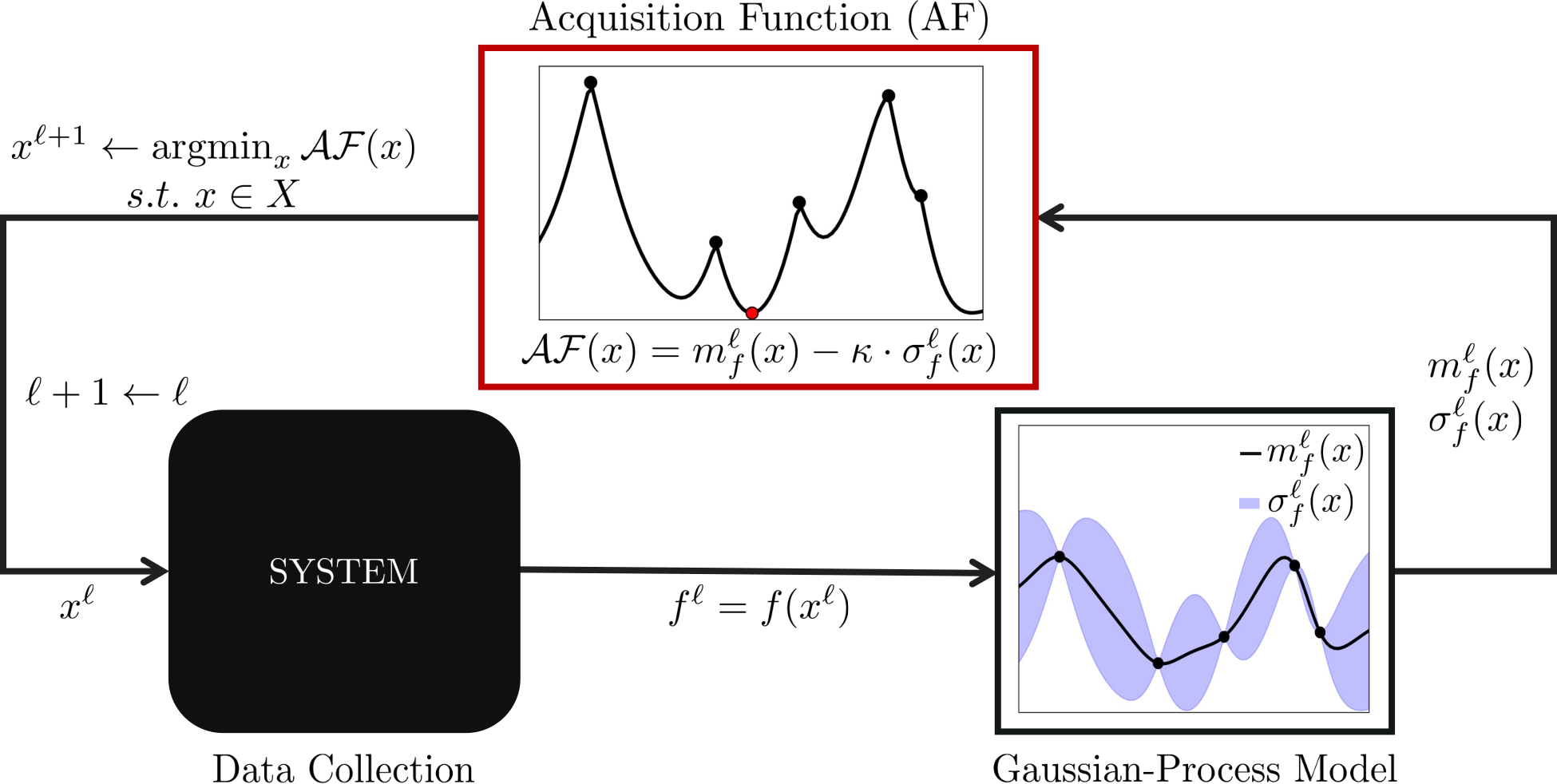}
	\caption{Block-flow diagram of the BO framework}
	\label{fig:S_BO}
\end{figure}
The problems we are interested in solving are of the form
\begin{subequations} \label{eq:goal}
    \begin{gather}
	\min_x~~f(x)\\
	\textrm{s.t.}~~x\in X
    \end{gather}
\end{subequations}
where $f: X\to \mathbb{R}$ is a scalar performance function, $X\subseteq\mathbb{R}^{d_x}$ is the design or search space, and $x$ is a set of design inputs within $X$. Generally, solving this problem is made difficult by the fact that derivatives cannot readily be calculated as sampling $f$ is expensive and the generated data can be noisy. Bayesian optimization manages these challenges by using input/output data to generate a surrogate model of the performance function that it uses to systematically explore the design space. The general framework for the algorithm is as follows: using an initial dataset of size ${\ell}$, $\mathcal{D}^{\ell} = \{x_\mathcal{K}, f_\mathcal{K}\}$, where $\mathcal{K}=\{1,...,\ell\}$, train a Gaussian process surrogate model. The GP assumes that the output data have a prior of the form $f(x_\mathcal{K})\sim\mathcal{N}\left(\textbf{m}(x),\textbf{K}(x, x^\prime)\right)$ where $\textbf{m}(x)\in\mathbb{R}^{d_x}$ is the mean function and $\textbf{K}(x, x^\prime)\in\mathbb{R}^{d_x\times d_x}$ is the covariance matrix. While $\textbf{m}(x)$ is usually set equal to \textbf{0}, $\textbf{K}(x, x^\prime)$ is calculated using a kernel function, $k(x, x^\prime)$, such that $\textbf{K}_{ij}=k(x_i, x_j)$. In our work, we have opted to use the M\'atern class of kernels to construct the covariance matrix. Once the GP has been conditioned on $\mathcal{D}^{\ell}$, it calculates the posterior distribution of $f$, $\hat{f}^{\ell}$, at a set of $n$ new points $\mathcal{X}$:
\begin{equation} \label{eq:posterior_dist}
	\hat{f}^{\ell}(\mathcal{X})\sim\mathcal{N}\left(m^{\ell}_f(\mathcal{X}), \Sigma^{\ell}_f(\mathcal{X})\right)
\end{equation}
where
\begin{subequations}\label{GP_moments}
    \begin{align}
	m_f^\ell(\mathcal{X}) & = \textbf{K}(\mathcal{X}, x_{\mathcal{K}})^T\textbf{K}(x_{\mathcal{K}}, x_{\mathcal{K}})^{-1}f_{\mathcal{K}}\\[5pt]
	\Sigma^{\ell}_f(\mathcal{X}) & = \textbf{K}(\mathcal{X}, \mathcal{X})-\textbf{K}(\mathcal{X}, x_{\mathcal{K}})^T\textbf{K}(x_{\mathcal{K}}, x_{\mathcal{K}})^{-1}\textbf{K}(x_{\mathcal{K}}, \mathcal{X})
    \end{align}
\end{subequations}
Equation \eqref{GP_moments} is used to construct an acquisition function (AF) of the form 
\begin{equation} \label{eq:acquisition_function}
	\mathcal{AF}(x)=m_f^\ell(x)-\kappa\cdot\sigma_f^\ell(x)
\end{equation}
that is then optimized to select a new sample point $x^{\ell+1}$. Note that the parameter $\kappa\in\mathbb{R}_{+}$, known as the exploration weight, determines the importance placed on the model uncertainty and can be modified to make the algorithm either more exploitative or explorative. After taking a sample at $x^{\ell+1}$, the dataset is updated and the model can be retrained. This process is repeated until a satisfactory solution is found or the data collection budget is exhausted. For a more complete treatment of BO we refer reader to \cite{Garnett:2023}.

\subsection{Monte Carlo approach for composite functions}
\begin{figure}[tb]
	\centering
	\includegraphics[width=0.45\textwidth]{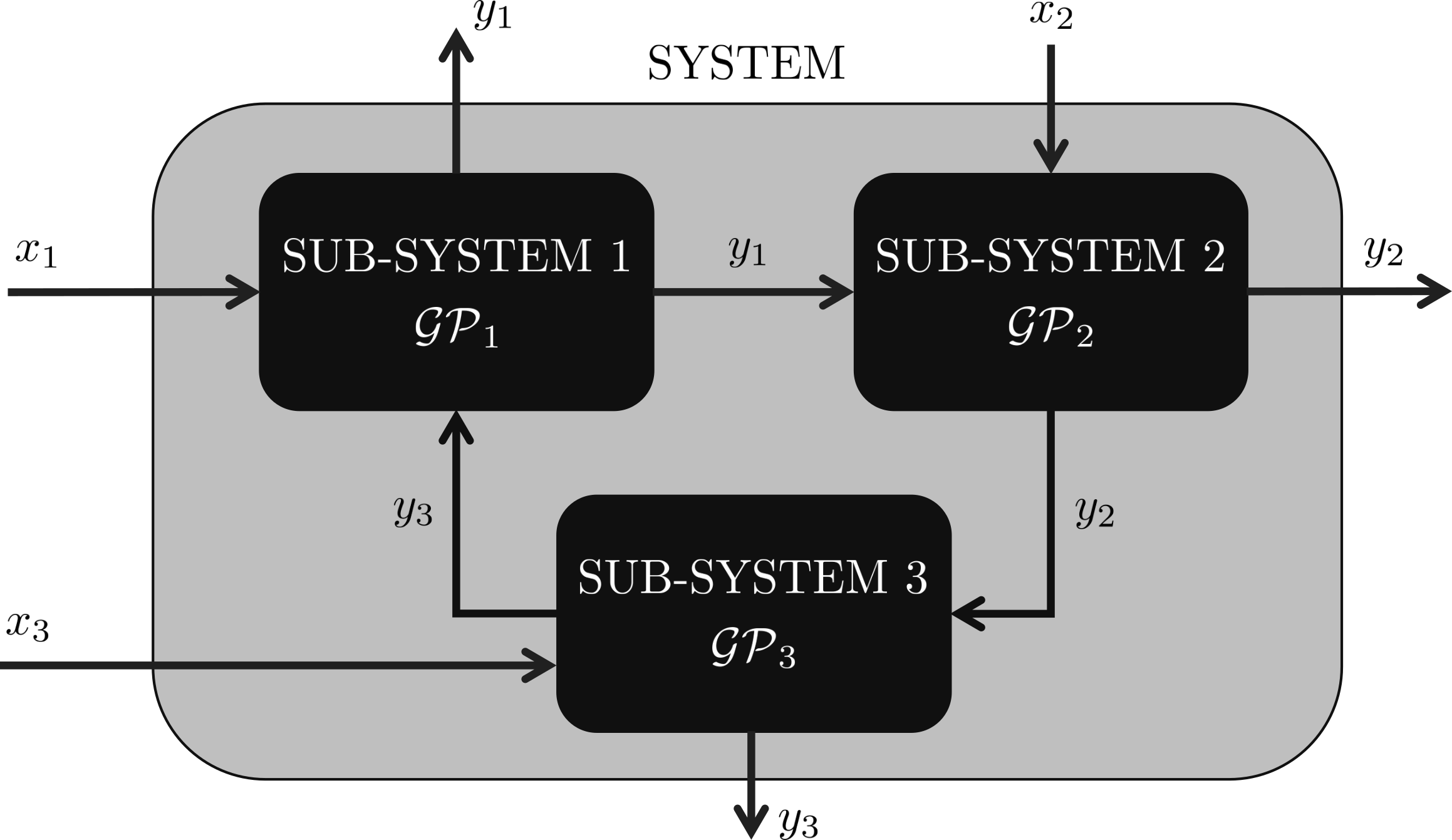}
	\caption{Grey-box representation of a composite function system with black-box intermediate functions}
	\label{fig:grey_box}
\end{figure}

Optimization of a composite objective using BO is introduced in \cite{Astudillo:2019}. In this context, the performance function is now a known composition of the form $f(x, y(x))$ with $f: X\times Y\to\mathbb{R}$. The inner or intermediate function $y: X\to \mathbb{R}^{d_y}$ is a vector-valued function with range $Y\subseteq\mathbb{R}^{d_y}$ and is now the unknown, expensive-to-evaluate function. Note that in this approach, the GP model no longer estimates the system's performance as in standard BO but is instead generating estimates for $y(x)$, which serve as inputs to the performance function. The system is then no longer a black-box but rather a composition of interconnected black-boxes whose relation to each other and contributions to the system's performance are known via $f(x, y(x))$. Additionally, because $f$ is a known function that can be easily evaluated, its derivatives are also available enabling the use of derivative-based techniques (gradient descent, Newton's method, etc.) to optimize the function. Thus, in this context, it would be be more precise to consider the system a partially observable "grey-box" rather than a black-box as shown in Figure \ref{fig:grey_box}.

In order to apply this paradigm in a BO setting, we must be able to provide the acquisition function with the mean and uncertainty estimates of $f$. However, because the performance function is no longer being approximated by the GP model, these are no longer as readily available as they were in the standard setting. Therefore, it is necessary to translate the mean and uncertainty estimates provided for $y$, $m_y^{\ell}(x)$ and $\Sigma_y^{\ell}(x)$ respectively, into the appropriate values for $m_f^{\ell}(x)$ and $\sigma_f^{\ell}(x)$. In the case where $f$ is a linear transformation of $y$ of the form $f = a^Ty+b$, then, given that the GP model assumes a normal distribution for $y$, $f$ is normally distributed with
\begin{subequations}\label{lin_f_moments}
    \begin{align}
	m_f^\ell(x) &  = a^Tm_y^{\ell}(x)+b\\[5pt]
	\sigma^{\ell}_f(x) & = \left(a^T\Sigma^{\ell}_y(x)a\right)^\frac{1}{2}
    \end{align}
\end{subequations}
However, in the more general case where $f$ is a nonlinear transformation, this property can no longer be used, and closed-form expressions for $m_f^{\ell}(x)$ and $\sigma_f^{\ell}(x)$ are not readily available. Various methods have proposed using Monte Carlo to address this problem, see \cite{Astudillo:2019}; \cite{Balandat:2020}; \cite{Paulson:2022}. Using this approach, the statistical moments of $f$ are estimated by passing samples from the GP models for $y(x)$ into $f(x, y(x))$ and then numerically estimating the mean and variance of the performance function as follows
\begin{subequations}\label{eq:MC_moments}
    \begin{align}
	\hat{m}_f^{\ell} & = \frac{1}{S}\sum_{s=1}^S f(x, m_y^{\ell}(x)+A_y^{\ell}(x)z_s)\\[5pt]
	\hat{\sigma}^{\ell}_f & = \frac{1}{S-1}\sqrt{\sum_{s=1}^S \left(f(x, m^{\ell}_y(x)+A_y^{\ell}(x)z_s)-\hat{m}^{\ell}_f\right)^2}
    \end{align}
\end{subequations}
where $A_y^{\ell}(x)\in\mathbb{R}^{d_y\times d_y}$ is the Cholesky factor of the GP covariance $\left(A_y^{\ell}(A_y^{\ell})^T=\Sigma_y^{\ell}\right)$ and $z\in\mathbb{R}^{d_y}$ is a random vector drawn from $\mathcal{N}(\textbf{0}, \textbf{I})$. These estimates are then passed into the the AF presented in \eqref{eq:acquisition_function}. 

While MC provides a convenient manner for determining the distribution of $f$, it is a computationally intensive method for doing so. The number of samples required to accurately estimate $m_f^{\ell}(x)$ and $\sigma_f^{\ell}(x)$ can be quite large (on the order of $10^3$ or more) in regions of the design space with high model uncertainty. As a result, drawing the number of samples, $S$, needed from a GP, which scales as $\mathcal{O}(S^3)$, see \cite{Shahriari:2016}, can require a significant amount of computational time. Additionally, while $f$ is a known function and is significantly cheaper to evaluate than the system, at large values of $S$ the cost of repeatedly calculating the value of $f(x, y(x))$ can also become nontrivial. This issue is compounded by the fact that \eqref{eq:MC_moments} must be recalculated at every point of interest.

\section{The BOIS Approach}

\begin{figure*}[!h]
	\centering
	\includegraphics[width=0.70\textwidth]{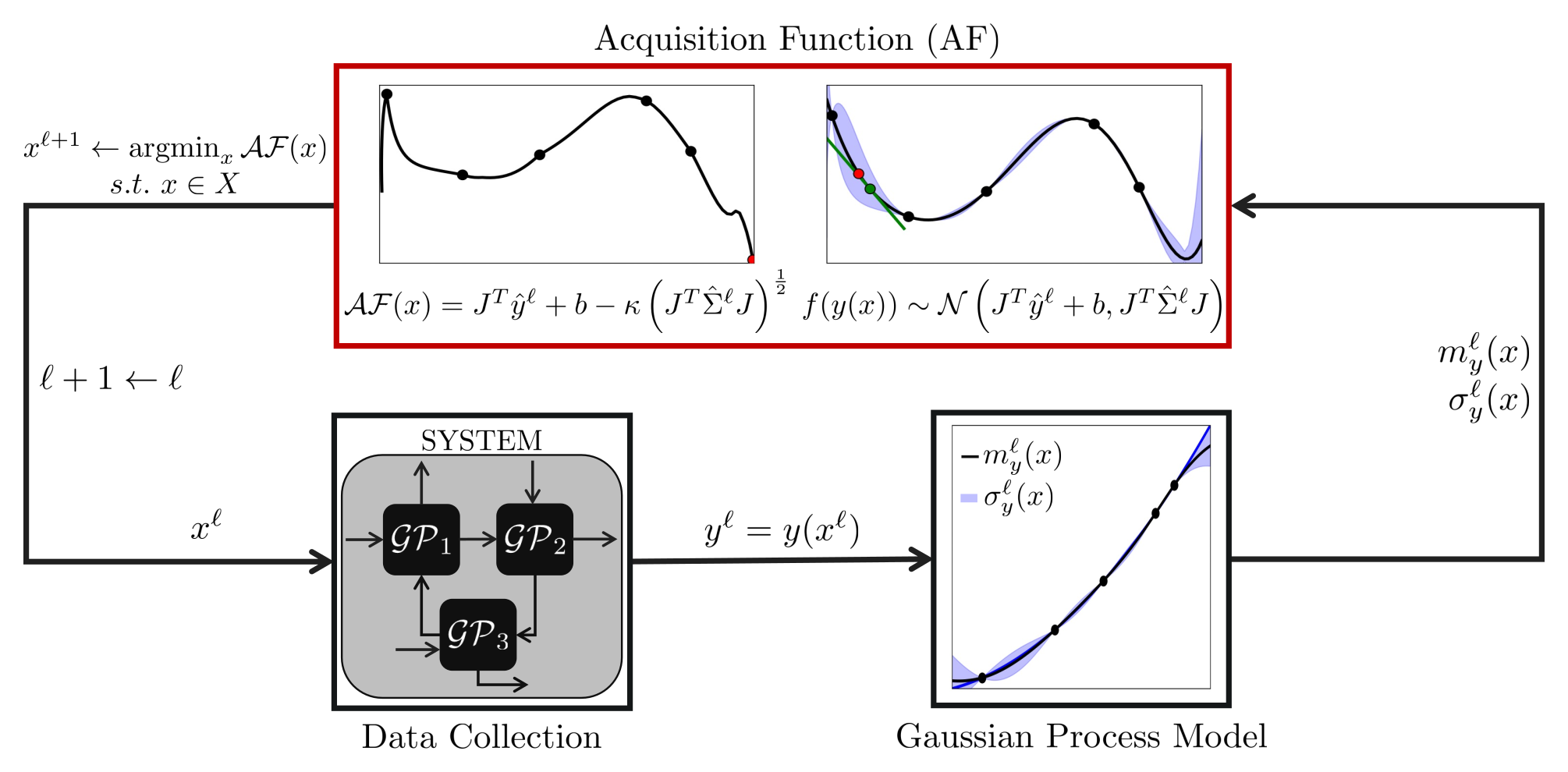}
	\caption{Representation of the BOIS framework, note that $b$ represents the deterministic terms in the linearized expression}
	\label{fig:BOIS}
\end{figure*}

The fundamental challenge of composite function BO algorithm is the lack of closed-form expressions for $m_f^{\ell}(x)$ and $\sigma_f^{\ell}(x)$, which are needed to build the AF. As previously mentioned, obtaining these when $f$ is not a linear mapping of $y$ is generally not possible. However, when $f$ is a once-differentiable mapping, it is possible to conduct a linearization at the current iterate (as is done in standard optimization algorithms such as Newton's method). If we choose to represent $f$ as
\begin{equation} \label{eq:split_f}
	f(x, y) = g(x)+h(x, y)
\end{equation}
such that $g(x)$ includes the terms that do \textit{not} depend on $y$ and $h(x ,y)$ includes those that do, we can use a first-order Taylor series expansion to linearize $f$ with respect to $y$ around some reference point $y_0$:
\begin{equation}\label{eq:linearization}
	f(x, y) \approx g(x)+h(x, y_0)+J^T(y-y_0)
\end{equation}  
where
\begin{subequations}\label{eq:gradient}
    \begin{align}
	J & = \nabla_yh(x, y_0) \\
	  & = \nabla_yf(x, y_0)
    \end{align}
\end{subequations}
Using this approach, if we then select some point of interest $x^{\ell+1}$ where the mean and covariance of $y(x)$ are given by the GP model as $\hat{y}^{\ell} = m_y^{\ell}(x^{\ell+1})$ and $\hat{\Sigma}^{\ell} = \Sigma_y^{\ell}(x^{\ell+1})$ respectively and some reference point in the neighborhood of $\hat{y}^{\ell}$, which we denote as $\hat{y}_0^{\ell}$, we can obtain the following estimate for $f$ at $x^{\ell+1}$
\begin{subequations} \label{eq:f_example}
    \begin{align}
	f(x^{\ell+1}, y(x^{\ell+1})) & \approx g(x^{\ell+1})+h(x^{\ell+1}, \hat{y_0}^{\ell}) \notag \\ & +J^T(y(x^{\ell+1})-\hat{y}_0^{\ell})
    \end{align}
\end{subequations}
Note that we make the assumption that $g(x)$ is also a known, easy-to-evaluate function and, therefore, $g(x^{\ell+1})$ is a \textit{deterministic} variable. Thus, we are now able to derive at a set of closed-form expressions for the mean and uncertainty of the performance function by making use of the closure of normal distributions under linear transformations.
\begin{subequations}\label{eq:lin_moments}
    \begin{align}
	m_f^{\ell}(x^{\ell+1}) & = J^T\hat{y}^{\ell}+g(x^{\ell+1})+h(x^{\ell+1},\hat{y}_0^{\ell})-J^T\hat{y}_0^{\ell}\\[5pt]
	\sigma^{\ell}_f(x^{\ell+1})& = \left(J^T\hat{\Sigma}^{\ell}J\right)^\frac{1}{2}.
    \end{align}
\end{subequations}
The proposed framework (which we call BOIS and is shown in Figure \ref{fig:BOIS}) is built on the expressions derived in \eqref{eq:lin_moments}. The reason we are able to generate these explicit formulations is due to the manner in which linearizing constructs the density of $f$. While the MC approach is agnostic to the nature of the density of $f$, BOIS approximates it using a Gaussian model around the neighborhood of the iterate $\hat{y}_0^{\ell}$. In other words, BOIS generates a local Laplace approximation of the performance function by passing the mean and uncertainty estimates of $y(x)$ given by the GP model into \eqref{eq:f_example}. This allows us to obtain closed-form expressions for the statistical moments of $f$, such as $m_f^{\ell}$ and $\sigma_f^{\ell}$, which can be used for constructing AFs. Given that this approximation of the density of $f$ is Gaussian, it is also possible to obtain expressions for the probabilities and quantiles (to construct different types of AFs). This assumption about the shape of $f$ also means that as the distance between $\hat{y}^{\ell}$ and $\hat{y}_0^{\ell}$ grows, the Laplace approximation will result in a worse fit, similar to how the linearization itself becomes less accurate. However, it is important to note that the linearization is updated in an adaptive manner (by linearizing around the current iterate). If we compare BOIS to MC-driven approaches, we can see that at any point $x^{\ell+1}$ of interest, BOIS only samples from the GP once to determine $\hat{y}^{\ell}$ and $\hat{\Sigma}^{\ell}$ and evaluates $f$ once to calculate $f(x^{\ell+1}, \hat{y}_0)$; recall that this is done tens to thousands of times in MC. While BOIS also has to compute \eqref{eq:gradient}, this is also done only once and has been shown to have a computational cost similar to that of evaluating $f$ when methods like automatic differentiation are used, see \cite{Griewank:2008}; \cite{Baur:1983}. As a result, we are able calculate values for $m_f^{\ell}$ and $\sigma_f^{\ell}$ at a significantly lower computational cost than when using MC.

\section{Numerical Experiments}

\subsection{Optimization of a chemical process}
\begin{figure*}[!htp]
	\centering
	\includegraphics[width=0.80\textwidth]{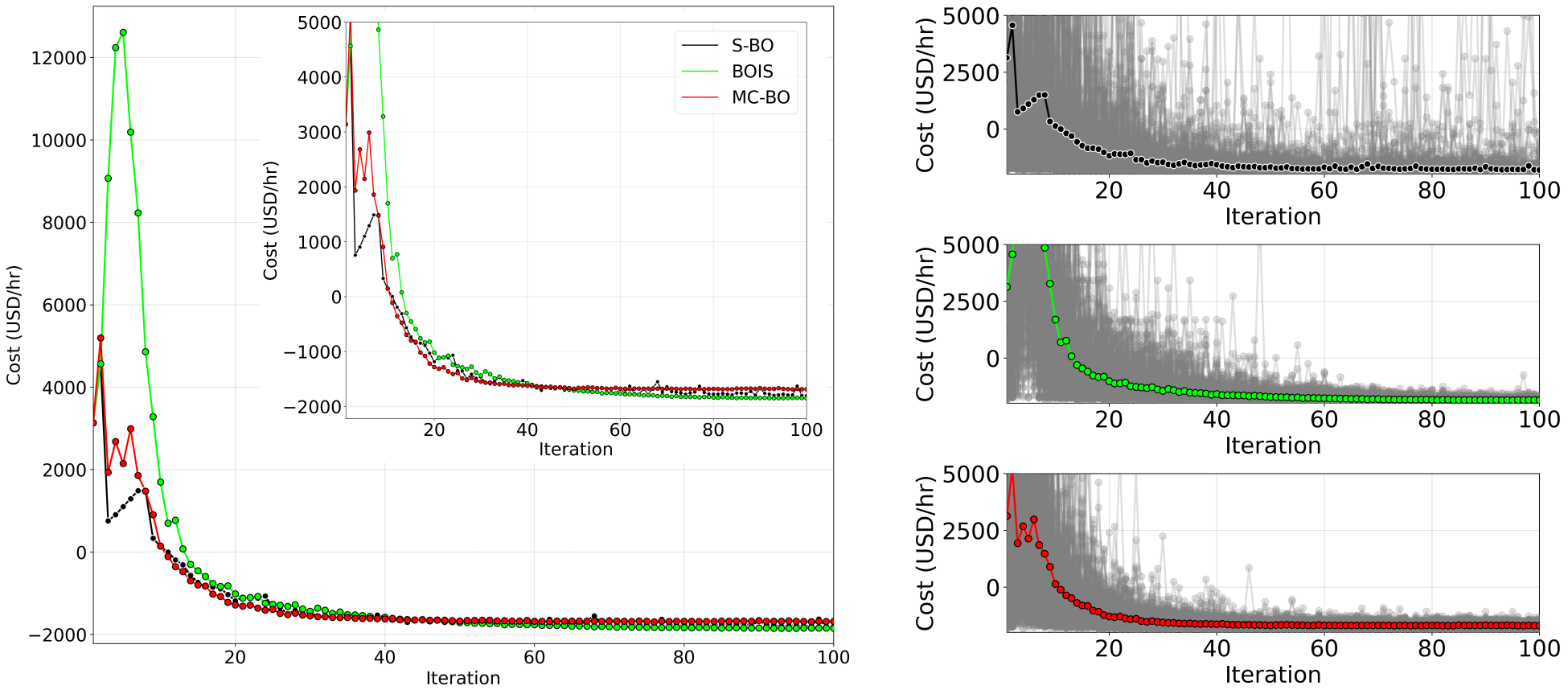}
	\caption{Iteration number vs average operating cost for the tested algorithms (left) and their distribution profiles across all runs with average performance shown in color (right)}
	\label{fig:results_1}
\end{figure*}
Consider the following chemical process: two reagents, $A$ and $B$, are compressed and heated and then fed into a reactor where they react to form product $C$. The reactor effluent is sent to a separator where the product is recovered as a liquid. A fraction of the vapor stream exiting the separator, which contains mostly unreacted $A$ and $B$, is recycled and fed back to the reactor after being heated and compressed, and the remainder is purged. Our aim is to determine the operating temperatures and pressures of the reactor and separator as well as the recycle fraction that will minimize the operating cost of the process. With this goal in mind, we formulate our cost function as
\begin{subequations}\label{eq:objective}
    \begin{align}
	f_1(y(x)) & = \sum_{j\in\{A,B\}}w_{j0}F_{j}+F_p\sum_{i\in\{A,B,C\}}w_{i1}\psi_i \notag \\ &+F_o\sum_{i\in\{A, B, C\}}w_{i2}\xi_i+w_3\left(\frac{\psi_CF_p-\bar{F}}{\bar{F}}\right)^2\\[5pt]
	f_2(y(x)) & = \sum_{h=1}^5 w_h\dot{Q}_h, +w_e\sum_{k=1}^3\dot{W}_k \\[5pt]
	f(y(x)) & = f_1(y(x))+f_2(y(x))
    \end{align}
\end{subequations} 
Here, $F_j$ are the flowrates of reagents into the process. The product and purge streams exit the process at rates $F_p$ with composition $\psi_i$ and $F_o$ with composition $\xi_i$ respectively. The heat and power requirements of the process units are denoted as $\dot{Q}_h$ and $\dot{W}_k$. The costs of reagents and heat and power utilities are $w_j$, $w_h$ and $w_e$ respectively while $w_{i1}$ and $w_{i2}$ refer to the values of species $i$ in the product and purge streams. We are also targeting a certain production rate for $C$, $\bar{F}$, which we choose to enforce by incurring an additional cost, $w_3$, when the process operates at a different value of $F_p$. We define our design space as the box domain $X = [673, 250, 288, 140, 0.5]\times[973, 450, 338, 170, 0.9]$; the optimal solution is at $x = (844, 346, 288, 170, 0.9)$. For the composite function BO algorithms, the intermediate black-box functions, $y(x)$, are set to model the flowrates and compositions of the product and purge streams as well as the heat and power loads of the process. All algorithms tested, standard BO (S-BO), MC-driven composite function BO (MC-BO), and BOIS, were initialized at the same point; we conducted 243 trials, each at a different starting point on a $3^5$ grid of $X$.

The left side of Figure \ref{fig:results_1} illustrates the average performance of the algorithms across all of the runs. We observe that, on average, BOIS is able to return a better solution than both S-BO and MC-BO. Additionally, we also see that during the first 10 trials BOIS samples from regions with significantly higher costs than the other two algorithms. We attribute this to the fact that, due to how $\sigma_f^\ell(x)$ is calculated in \eqref{eq:lin_moments}, large values in $J$ and $\Sigma_y^{\ell}(x)$ can result in the uncertainty estimate for $f(y(x))$ being high. This pushes the algorithm to be more explorative, especially at the beginning of the run when the model uncertainty of $y(x)$ is at its highest. From the sharpness of this peak we can also determine that BOIS quickly moves away from these highly non-optimal areas; after 15 iterations it is exploring regions with similar values as BO and MC-BO.  

The performance distributions shown on the right side of Figure \ref{fig:results_1} clearly illustrate the benefit of using a composite representation of the performance function. We observe that while the sampling behavior of BOIS and MC-BO is relatively stable after approximately 60 iterations across all runs, several instances of S-BO are still sampling from high-cost regions even at the end of their runs. The cause of this is the flow penalty term included in \eqref{eq:objective}. By providing the composite function algorithms with the form of $f$, we enable them to easily determine values of $F_p$ that result in a small penalty. S-BO does not have access to this information and, as a result, has no way of knowing that the penalty is there and has to rely on sampling to learn it. From the results illustrated in the figure, we can surmise that this is a difficult task for the algorithm. 

\subsection{Statistical consistency of BOIS}
\begin{figure*}[!h]
	\centering
	\includegraphics[width=0.80\textwidth]{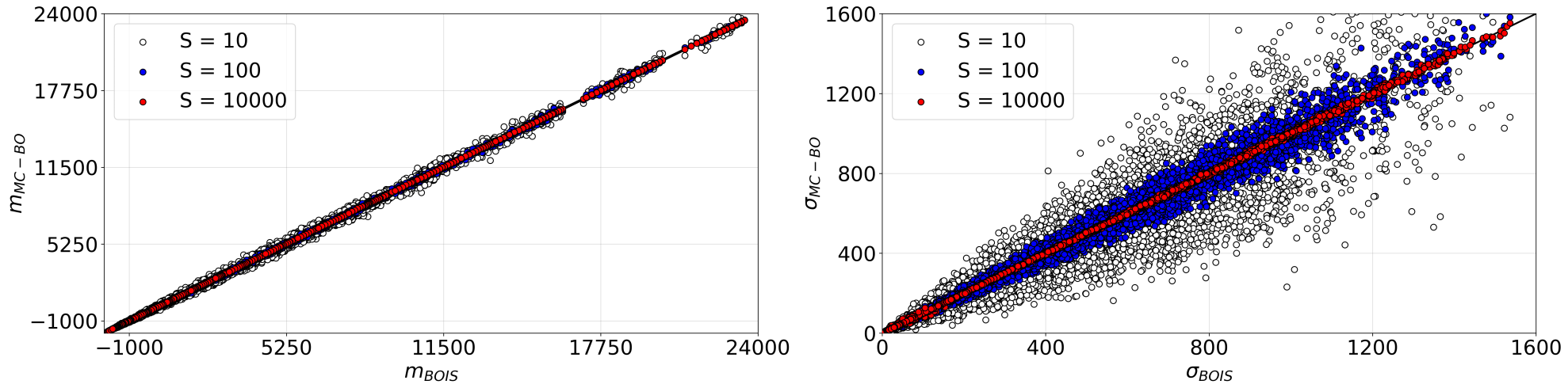}
	\caption{Parity plots of the values of $m_f^{\ell}$ and $\sigma_f^{\ell}$ at various points in $X$ calculated by BOIS with $\hat{y}-\hat{y}_0=\hat{y}\times 10^{-3}$ and MC-BO using samples sizes $S = 10$, $10^2$, and $10^4$; the same trained GP model of $y(x)$ was used by both approaches}
	\label{fig:results_2}
\end{figure*}
We decided to use the accuracy of the statistical moments of $f$ calculated by BOIS as a metric for comparing its efficacy with that of MC-BO. We know that as we increase the number of samples \eqref{eq:MC_moments} will return values closer to the true moments of $f$. Using a trained GP model of $y(x)$ we generated estimates for $m_f^{\ell}$ and $\sigma_f^{\ell}$ at various points in the design space using MC-BO with 10, 100, and 10,000 samples. We then used this same GP model and set $\hat{y}-\hat{y}_0=\hat{y}\times 10^{-3}$ in the linearization step to obtain the corresponding estimates from BOIS. If the values given by BOIS are accurate, then we should expect that the difference between its estimates and those of MC-BO should decrease as $S$ increases. The results presented in Figure \ref{fig:results_2} demonstrate that this precisely the case. The shift in the difference between the estimates is especially clear when looking at the values of $\sigma_f^{\ell}$ calculated by the two algorithms. The large discrepancies seen when using 10 samples shrink significantly when $S$ is increased to 100 and are virtually gone when $S=$ 10,000. If we consider the time it took to generate these estimates, approximately 10 seconds for BOIS and 1 hour for MC-BO when using 10,000 samples, we can conclude that, not only is BOIS faster than MC-BO, it can also be just as accurate. This further reinforces our claim that BOIS is an efficient method for using composite functions in a BO setting.

\section{Conclusions}
We presented a framework, which we refer to as BOIS, to enable the use of composite functions $f(x,y(x))$ in BO to allow for the exploitation of structural knowledge (in the form of physics or sparse interconnections). The key contribution of this work is the derivation of a set of closed-form expressions for the moments of the composite, $m_f^{\ell}(x)$ and $\sigma_f^{\ell}(x)$, based on adaptive linearizations; this overcomes the tractability challenges of sampling based approaches. These expressions are obtained by linearizing $f(x, y(x))$ in order to generate a Laplace approximation around the current iterate. We demonstrate that BOIS outperforms standard BO by making use of the structure conveyed by $f(x, y(x))$. We also show that the statistical moments calculated by BOIS accurately represent the statistical moments of $f$ and that these estimates can be obtained in significantly less time than sampling-based approaches. As part of our future work, we plan to scale up BOIS and deploy it on high-dimensional systems where BO has traditionally not been applied and to obtain alternative types of AFs. Finally, while the GP is the most popular surrogate model choice in BO, the algorithm is not limited to using only GPs, any probabilistic model can be used. Therefore, we would like to explore the use of alternative models such as warped GPs, see \cite{Snelson:2004}, RNNs, see \cite{Thompson:2023}, and reference models, see \cite{Lu:2021}. 

\bibliography{root}             % bib file to produce the bibliography
                                                     % with bibtex (preferred)
                                                   
%\begin{thebibliography}{xx}  % you can also add the bibliography by hand

%\bibitem[Able(1956)]{Abl:56}
%B.C. Able.
%\newblock Nucleic acid content of microscope.
%\newblock \emph{Nature}, 135:\penalty0 7--9, 1956.

%\bibitem[Able et~al.(1954)Able, Tagg, and Rush]{AbTaRu:54}
%B.C. Able, R.A. Tagg, and M.~Rush.
%\newblock Enzyme-catalyzed cellular transanimations.
%\newblock In A.F. Round, editor, \emph{Advances in Enzymology}, volume~2, pages
%  125--247. Academic Press, New York, 3rd edition, 1954.

%\bibitem[Keohane(1958)]{Keo:58}
%\newblock \emph{Power and Interdependence: World Politics in Transitions}.
%\newblock Little, Brown \& Co., Boston, 1958.

%\bibitem[Powers(1985)]{Pow:85}
%\newblock Is there a way out?
%\newblock \emph{Harpers}, pages 35--47, June 1985.

%\bibitem[Soukhanov(1992)]{Heritage:92}
%\newblock \emph{{The American Heritage. Dictionary of the American Language}}.
%\newblock Houghton Mifflin Company, 1992.

%\end{thebibliography}
\end{document}